%% file: main.tex
\documentclass{article}
\usepackage{amsmath,graphicx}
\usepackage{amsmath}
\usepackage{cases}
\usepackage{mathtools}
\usepackage{amssymb}
\usepackage{amsthm}
\usepackage{bm}
\usepackage{algorithm}
\usepackage{algpseudocode}
\usepackage{hyperref}
\usepackage{gensymb}
\usepackage{placeins}
\usepackage[dvipsnames]{xcolor}
\usepackage{tikz}
\usetikzlibrary{backgrounds}
\usetikzlibrary{arrows,shapes}
\usetikzlibrary{tikzmark}
\usetikzlibrary{calc}
\usepackage{mathtools, nccmath}
\usepackage{wrapfig}
\usepackage{comment}
\usepackage{tcolorbox}
\usepackage{tikz}
\usetikzlibrary{arrows,shapes,positioning,shadows,trees,mindmap}
\usepackage[edges]{forest}
\usetikzlibrary{arrows.meta}
\colorlet{linecol}{black!75}
\usepackage{xkcdcolors} 
\usepackage{tikz}
\usetikzlibrary{backgrounds}
\usetikzlibrary{arrows,shapes}
\usetikzlibrary{tikzmark}
\usetikzlibrary{calc}
\usepackage{authblk}
\newcommand{\highlight}[2]{\colorbox{#1!17}{$\displaystyle #2$}}

\colorlet{mhpurple}{Plum!80}
\renewcommand{\highlight}[2]{\colorbox{#1!17}{#2}}

\newcommand{\norm}[1]{\left\lVert#1\right\rVert}
\newfloat{prob}{thp}{lop}
\floatname{prob}{Problem}
\newcommand\scalemath[2]{\scalebox{#1}{\mbox{\ensuremath{\displaystyle #2}}}}
\title{Globally Optimal Boresight Alignment of UAV-LiDAR Systems}
\author[$\dagger$]{Smitha Gopinath, Hassan L. Hijazi, Adam Collins, Julian Dann Nathan Lemons, Emily Schultz-Fellenz, Russell Bent}
\author[$\star$]{Amira Hijazi}
\author[$\ddag$]{Gert Riemersma}
\affil[$\dagger$]{Los Alamos National Laboratory}
\affil[$\star$]{North Carolina State University}
\affil[$\ddag$]{Routescene Inc. Mapix Technologies Ltd}

\begin{document}
%
\maketitle

\insert\footins{
  \normalfont\footnotesize
  \interlinepenalty\interfootnotelinepenalty
  \splittopskip\footnotesep \splitmaxdepth \dp\strutbox
  \floatingpenalty10000 \hsize\columnwidth
  \copyright 20XX IEEE. Personal use of this material is permitted. Permission from IEEE must be obtained for all other uses, in any current or future media, including reprinting/republishing this material for advertising or promotional purposes, creating new collective works, for resale or redistribution to servers or lists, or reuse of any copyrighted component of this work in other works.}
  
\begin{abstract}
In airborne light detection and ranging (LiDAR) systems, misalignments between the LiDAR-scanner and the inertial navigation system (INS) mounted on an unmanned aerial vehicle (UAV)'s frame can lead to inaccurate 3D point clouds. Determining the orientation-offset, or boresight error, is key to many LiDAR-based applications. In this work, we introduce a mixed-integer quadratically constrained quadratic program (MIQCQP) that can globally solve this misalignment problem. We also propose a nested spatial branch and bound (nsBB) algorithm that improves computational performance. The nsBB relies on novel preprocessing steps that progressively reduce the problem size. In addition, an adaptive grid search (aGS) allowing us to obtain quick heuristic solutions is presented. Our algorithms are open-source, multi-threaded and multi-machine compatible.
\end{abstract}
%
%
\newpage
\section{Introduction}
\label{sec:intro}

LiDAR-scanners generate high-density, three-dimensional point clouds that may be used to either build topographic maps or digital elevation models. Airborne LiDAR-based sensing and mapping has several uses ranging from urban planning \cite{urban} and infrastructure monitoring \cite{bridge} to disaster response \cite{disaster}. Georeferencing of LiDAR data is essential for mapping, data visualization (from multiple scans of an area) and sensor fusion.

Georeferencing consists of expressing the coordinates of each point $p_i$ in a given (fixed) coordinate system called the mapping frame \cite{Schenk}:
\vspace{\baselineskip}
\begin{equation}
\vspace{\baselineskip}
p_i=\tikzmarknode{s}{\highlight{blue}{$\bm{s}_i$}}+\tikzmarknode{i}{\highlight{blue}{$\bm{R}_i$}}\tikzmarknode{r}{\highlight{red}{$R_{\alpha\beta\gamma}$}}\tikzmarknode{l}{\highlight{blue}{$\bm{l}_i$}} \label{eq:geo}
\end{equation}
\begin{tikzpicture}[overlay,remember picture,>=stealth,nodes={align=left,inner ysep=1pt},<-]

    \path (s.north) ++ (0,1em) node[anchor=south east,color=blue!67] (scalep){{scanner position}};
    \draw [color=blue!67](s.north) |- ([xshift=-0.3ex,color=red]scalep.south west);

    \path (i.south) ++ (0,-1.6em) node[anchor=south east,color=blue!67] (mean){INS rotation};
   \draw [color=blue!67](i.south) |- ([xshift=-0.5ex,color=red]mean.south west);
   
       \path (r.south) ++ (0,-1.6em) node[anchor=south west,color=red!67] (bore){Boresight error};
   \draw [color=red!67](r.south) |- ([xshift=+0.3ex,color=red]bore.south east);
   
   \path (l.north) ++ (0,1em) node[anchor=south,color=blue!67] (lid){{point $i$ w.r.t. LiDAR-scanner}};
    \draw [color=blue!67](l.north) |- ([xshift=+0.1ex,color=blue]lid.south east);

\end{tikzpicture}
where, $\bm{l}_i$ is the coordinates of point $i$ in the LiDAR-scanner frame, $\bm{R}_i \in  {\cal SO}(3)$ is the rotation of the INS frame w.r.t. the mapping frame, $\bm{s}_i$ is the position of the LiDAR-scanner in the mapping-frame ($\bm{s}_i$ is the sum of the INS-sensor position and the scanner position-offset in the mapping-frame), and $R_{\alpha\beta\gamma} \in  {\cal SO}(3)$ is the rotation of the LiDAR-scanner frame w.r.t to the INS frame.

The misalignment between the sensors $R_{\alpha\beta\gamma}$ is also referred to as mounting-bias or boresight error. $R_{\alpha\beta\gamma}$ can vary each time the sensors are mounted on to the UAV (due to mechanical stresses or human error) \cite{Katzenbeisser} \cite{three_deg}. Even though these angles are usually small (less than 3\degree \cite{three_deg}), their correct estimation is key to accuracy of point clouds: a 1\degree error in the boresight roll may cause position errors greater than 5 m \cite{5merror}. The nature of LiDAR data presents a challenge to boresight alignment: LiDAR scanners generate a high number (in the order of millions) of unordered, discrete points with nonuniform point-density \cite{primitives_review}, unlike, say, RGB images. State-of-the-art image-based algorithms often fail on LiDAR data \cite{primitives_review}, necessitating new LiDAR-centric algorithms and tools.

Most boresight alignment tools available today are commercial and only compatible with a limited set of LiDAR and INS hardware, with the exception of the open-source project OpenMMS~\cite{openmms}. To the best of our knowledge, all existing algorithms for the boresight alignment problem fall under heuristics with no optimality guarantees. \cite{Skaloud} rely on manual identification of planar features in the point-cloud, whereas, \cite{keyetieu2019automatic} automate this. However, planar features may be absent in non-urban settings. Other methods introduce artificial control points \cite{ground_control} and reflective targets \cite{reflective_targets} in the scene. Others rely on matching points \cite{approx_icp}, profiles \cite{raviprofile} and features \cite{Habib2005} in two overlapping LiDAR scans. Variants of the iterative closest point (ICP) \cite{besl1067method} heuristic have also been used in several studies \cite{approx_icp,objectsegm}. It is worth noting that the boresight alignment problem bears a resemblance to the classic point set registration problem \cite{goicp,Teaser}.
This being said, registration tools fail to perform boresight alignment due to one fundamental dissimilarity: while registration algorithms assume a fixed point of reference for all rotations, this point is time-varying for boresight alignment (as can be seen in \eqref{eq:geo}), rotations are relative to the moving UAV frame.

At the heart of the boresight alignment problem is the solution of a nonlinear least squares problem to find the rotation matrix $R_{\alpha\beta\gamma}$. Previously developed approaches may converge to incorrect local minima \cite{global} as rotation matrices form a non-convex set. In most studies, the rotation matrix is linearized \cite{Biaspaper} which may lead to sub-optimal solutions. Global optimality guarantees are essential to benchmarking and evaluating heuristics, and perhaps training machine learning models.

\subsection{Contributions}
Here, we present the first open-source global optimization tool for boresight alignment. Our method is independent of hardware and simply uses LAZ \cite{LASlib} files generated from any UAV+INS+LiDAR system. The LAZ file has the point coordinates (in scanner frame) as well as the orientation and position of the INS w.r.t. the mapping frame which we assume to be accurate.

We use data-sets $\cal \hat{{\cal D}}$ and $\bar{{\cal D}}$ generated from overlapping scans of two parallel flight lines in opposite directions.
We may use any small subset of the data (of the order of thousand points) for calibration. As data from planes with different spatial orientations is needed to estimate all three boresight angles \cite{Skaloud}, we focus on a 3D object present in both sets (similar to \cite{objectsegm}). We optimize over data-sets  $\cal \hat{{\cal P}}$ and $\bar{{\cal P}}$, ($|\cal \hat{{\cal P}}| \le |\bar{{\cal P}}|$).

Our main contributions to the boresight alignment problem are as follows:
i) A new MIQCQP formulation of the problem;
ii) A nested spatial branch and bound algorithm which solves the problem to global optimality;
iii) A fast adaptive grid search heuristic.
\section{Mathematical Formulation}\label{sec:math}
\subsection{Rotation}
Any 3 $\times$ 3 matrix in the set of rotations ${\cal SO}(3)$ is parameterized by angles $\alpha,\, \beta,\, \gamma$:
\begin{align}
&R_{(\alpha\beta\gamma)}=\nonumber \\
& \scalemath{0.60}{ \begin{bmatrix}
& \cos (\beta) \cos (\gamma) & -\cos (\beta) \sin (\gamma) & \sin (\beta)\\
&  \cos (\alpha)\sin (\gamma)+\sin (\alpha)\sin (\beta)\cos (\gamma) &  \cos (\alpha)\cos (\gamma)-\sin (\alpha)\sin (\beta) \sin (\gamma) &  -\cos (\beta)\sin (\alpha)\\
& \sin (\alpha)\sin (\gamma)-\cos (\alpha)\sin (\beta)\cos (\gamma) & \sin (\alpha)\cos (\gamma)+\cos (\alpha)\sin (\beta)\sin (\gamma) & \cos (\alpha) \cos (\beta)
\end{bmatrix}} \label{eq:rot}
\end{align}
In this work, we use an equivalent quadratic formulation of \eqref{eq:rot} defined below:
\begin{align*}
& u_i^2+v_i^2=1 \quad \forall i \in \{\alpha,\beta,\gamma\},w_{\gamma\beta}=u_{\gamma}v_{\beta}, w_{\beta\gamma}=v_{\beta}v_{\gamma}\\
& \scalemath{0.96}{{R_{(\alpha\beta\gamma)} = \begin{bmatrix} \label{eq:r2} \tag{Q}
u_{\beta}u_{\gamma} & -u_{\beta}v_{\gamma} & v_{\beta}\\  
u_{\alpha}v_{\gamma}+v_{\alpha}w_{\gamma\beta} & u_{\alpha}u_{\gamma}-v_{\alpha}w_{\beta\gamma} & -u_{\beta}v_{\alpha}\\
v_{\alpha}v_{\gamma}-u_{\alpha}w_{\gamma\beta} & v_{\alpha}u_{\gamma}+u_{\alpha}w_{\beta\gamma} & u_{\alpha}u_{\beta} 
\end{bmatrix}}}
\end{align*}

\subsection{Formulation of the Boresight Alignment problem}
The Alignment problem is given in \ref{p:summin}. Each $i \in \hat{{\cal P}}$ is matched with a $j \in \bar{{\cal P}}$, such that the squared L2 distance between these points is the closest. We choose a rotation $R_{\alpha\beta\gamma}$ such that the total squared L2 error is minimized.

\begin{prob}
\caption{Boresight Alignment}\label{p:summin}{
\begin{align}
& \min_{\alpha,\beta,\gamma}  \sum_{i \in \hat{{\cal P}}}\min_{j \in \bar{{\cal P}}}\norm{(\hat{\bm{s}}_i    + \hat{\bm{R}}_i R_{\alpha\beta\gamma} \hat{\bm{l}}_i)-(\bar{\bm{s}}_j   + \bar{\bm{R}}_j R_{\alpha\beta\gamma}\bar{\bm{l}}_j)}_2^2 \nonumber\\
 & (\alpha,\beta,\gamma) \in [\alpha^{L},\alpha^{U}]\times[\beta^{L},\beta^{U}]\times[\gamma^{L},\gamma^{U}] \nonumber
\end{align}
}
\end{prob}

To express problem \ref{p:summin} as a mathematical program, we introduce binary variables $b_{i,j}$ to select a point in set $\bar{{\cal P}}$ (see \eqref{a:lmn} and  \eqref{a:bij}) that is closest to $i$ when the two point-clouds are transformed via rotation $R_{\alpha\beta\gamma}$ (\eqref{a:pe} and \eqref{a:pf}). Bounds on $u_i,v_i$ are derived from bounds on $\alpha,\beta,\gamma$. 
\begin{prob}
\caption{MIQCQP formulation of Boresight Alignment}\label{p:Align}{
\begin{subequations}
\begin{align}
& \min_{\substack{u_\alpha, u_\beta, u_\gamma\\ v_\alpha, v_\beta,v_\gamma \\ w_{\gamma\beta},w_{\beta\gamma}, b}}  \sum_{i \in \hat{{\cal P}}} \norm{\hat{p}_i-p_i}_2^2 \label{b:obj} \\
& p_i=\sum_{j: (i,j) \in  {\cal B}} \bar{p}_j b_{i,j} ~~, \forall i \in \hat{{\cal P}} \label{a:lmn} \\
& \hat{p}_i = \hat{\bm{s}}_i    + \hat{\bm{R}}_i R_{\alpha\beta\gamma} \hat{\bm{l}}_i ~~, \forall i \in \hat{{\cal P}} \label{a:pe}\\
& \bar{p}_j = \bar{\bm{s}}_j   + \bar{\bm{R}}_j R_{\alpha\beta\gamma}\bar{\bm{l}}_j ~~, \forall j \in \bar{{\cal P}} \label{a:pf} \\
& \sum_{j: (i,j) \in  {\cal B} } b_{i,j}=1 ~~, \forall i \in  \hat{{\cal P}}  \label{a:bij}\\
& \eqref{eq:r2}\nonumber\\
& b \in \{0, 1 \}^{|{\cal B}|}  \label{a:bin}\\
& u^L_i \le u_i \le u^U_i, v^L_i \le v_i \le v^U_i  ~~, \forall i \in \{\alpha,\beta,\gamma\}
\end{align}
\end{subequations}
}
\end{prob}\vspace*{-0.5cm}

\section{Model-size Reduction}
As point-correspondences are unknown, ${\cal B}$ in \ref{p:Align} allows each point in $\hat{{\cal P}}$ to be paired with every point in $\bar{{\cal P}}$. Thus, $|{\cal B}|=|\cal \hat{{\cal P}}| \times |\bar{{\cal P}}|$. However, the large number of binary variables makes the solution of MIQCQP \ref{p:Align} intractable by state-of-the-art MIP solvers. In this work, we reduce the size of the set  ${\cal B}$, by reasoning about the optimal solution of \ref{p:Align} (denoted by $^*$). 

Let $f^U$ be a valid upper-bound on the objective-value of \ref{p:Align}, such that $f^* \le f^U$. The objective function of problem \eqref{p:Align} may be equivalently expressed as $\sum_{(i,j) \in  {\cal B}} \norm{\hat{p}_i-\bar{p}_j}_2^2 b_{i,j}$, a weighted sum of pair-wise squared distances for every pair $(i,j) \in {\cal B}$. If we can prove apriori that $b^*_{i,j}=0$, then we may remove pair $(i,j)$ from the set ${\cal B}$ without affecting the optimal solution of problem \ref{p:Align}. To do so, we compute lower and upper bounds on the distance between each pair of points.

Let $c^L_{i,j}$ be a lower bound on the squared distance between a point $\hat{p}_i$ and $\bar{p}_j$. $c^L_{i,j} \le d^L_{i,j}$ where,
\begin{align*}
 d^L_{i,j} = &\min \norm{\hat{p}_i-\bar{p}_j}_2^2 \\
& \hat{p}_i = \hat{\bm{s}}_i    + \hat{\bm{R}}_i { R_{\alpha\beta\gamma} \hat{\bm{l}}_i} \\
& \bar{p}_j = \bar{\bm{s}}_j   + \bar{\bm{R}}_j  { R_{\alpha\beta\gamma}\bar{\bm{l}}_j } \label{prob:dl} \tag{DL}\\
& R_{\alpha\beta\gamma} \in {\cal SO}(3)\\
& \alpha,\beta,\gamma \in [\alpha^{L},\alpha^{U}],\, [\beta^{L},\beta^{U}],\,[\gamma^{L},\gamma^{U}].
\end{align*}
If $c^L_{i,j}>f^U$, then $b^*_{i,j}=0$ (since $f^* \le f^U$). Hence, $(i,j)$ may be removed from the pair-set ${\cal B}$ without affecting the solution of problem \ref{p:Align}.

Analogously, $c^U_{i,j}$ is an upper-bound on the squared distance between a point $\hat{p}_i$ and $\bar{p}_j$.
$b^*_{i,k}=1$ at the optimum if and only if,
$
\begin{aligned}
\norm{\hat{p}_i-\bar{p}_k}_2^2 \le \norm{\hat{p}_i-\bar{p}_j}_2^2 \quad \forall j \in \bar{{\cal P}}\end{aligned}$.\\ That is, point $i$ is paired with the closest point $k \in \bar{{\cal P}}$. For any $j,k \in \bar{{\cal P}}$, if $c^U_{i,k} < c^L_{i,j}$, (since $\bar{p}_j$ is not the closest point to $\hat{p}_i$) we certify $b^*_{i,j}=0$ and ${\cal B}={\cal B}\setminus (i,j)$ 

\subsection{Computing bounds on pair-wise distances}
The feasible regions for problem \eqref{prob:dl} is non-convex. In this section, we develop a tight linear relaxation of the feasible region in order to make the bounding problems tractable.  We use $\textrm{conv}(X)$ to denote the convex-hull of a set $X$ and $\textrm{V}(X)$ to denote the set of vertices of a polytope $X$.

Consider the set ${\cal X}(\bm{s}_i, \bm{R}_i, \bm{l}_i) \subset \mathbb{R}^3$ and its convex relaxation ${\cal C}(\bm{s}_i, \bm{R}_i, \bm{l}_i) \subset \mathbb{R}^3$.
\begin{align*}
\scalemath{0.92}{
    {\cal X}(\bm{s}_i, \bm{R}_i, \bm{l}_i)=x: \left \{ \begin{aligned}  &x=\bm{s}_i+ \bm{R}_i R \bm{l}_i\\ & R_{\alpha\beta\gamma} \in  {\cal SO}(3) \\ 
    & \alpha,\beta,\gamma \in [\alpha^{L},\alpha^{U}],[\beta^{L},\beta^{U}],[\gamma^{L},\gamma^{U}] \end{aligned} \label{setx} \tag{X} \right \}
    }
\end{align*}

We first examine the set ${\cal X}(0, \textrm{I}, \bm{l}_i)$, where $\textrm{I}$ denotes the 3 $\times$ 3 identity matrix. By domain-propagation of constraints in \eqref{setx}, we find bounded box $K \equiv x^L \le x \le x^U$. ${\cal X}(0, \textrm{I}, \bm{l}_i) \subset K$. Further, we compute the norm of $x \in {\cal X}(0, \textrm{I}, \bm{l}_i)$. Since $R^TR=\textrm{I}$ for any rotation matrix, we get,
\begin{align*}
    \norm{x}_2^2=&\norm{0+\textrm{I}R_{\alpha\beta\gamma}\bm{l}_i}_2^2
    =&\bm{l}^T_iR^T_{\alpha\beta\gamma}R_{\alpha\beta\gamma}\bm{l}_i
    =\norm{\bm{l}_i}_2^2.
\end{align*}
Thus we get the following equation,
\begin{align}
& \sum^3_{e=1} x_e^2 = \norm{\bm l_i}_2^2 \quad \forall x \in {\cal X}(0, \textrm{I}, \bm{l}_i) \label{eq:vi}
\end{align}
\eqref{eq:vi} may be convexified via a tangent \eqref{eq:tangent} at any $x^k \in K$, and using \eqref{eq:secant} as a concave envelope.
\begin{align}
& \sum^3_{e=1} (2x_e^kx_e -{x_e^k}^2) \le \norm{\bm l_i}_2^2 && \forall x \in {\cal C}(0, \textrm{I}, \bm{l}_i) \label{eq:tangent}\\
& \norm{\bm{l}_i}_2^2 \le \sum^3_{e=1} (x_e(x^L_e+x^U_e)-x^L_e x^U_e) && \forall x \in {\cal C}(0, \textrm{I}, \bm{l}_i) \label{eq:secant}
\end{align}

 ${\cal C}(0, \textrm{I}, \bm{l}_i)$ is given by intersection of $K$ with \eqref{eq:tangent} and \eqref{eq:secant}. $\textrm{V}({\cal C}(0, \textrm{I}, \bm{l}_i)$ is found by either vertex-enumeration or linear-algebra. As ${\cal C}(\bm{s}_i, \bm{R}_i, \bm{l}_i)$ is the result of an affine transformation of ${\cal C}(0, \textrm{I}, \bm{l}_i)$, we get,
\begin{align}
\textrm{V}({\cal C}(\bm{s}_i, \bm{R}_i, \bm{l}_i))=\{\bm{s}_i+\bm{R}_iv:  v \in \textrm{V}({\cal C}(0, \textrm{I}, \bm{l}_i)) \}.\label{eq:sety}
\end{align}
Given $\textrm{V}({\cal C})$, $c^L_{i,j}$ may be computed by convex optimization of \eqref{CL}. Here, we use a fast routine dedicated to distance minimization, openGJK \cite{openGJK}. To compute $c^U_{i,j}$, we enumerate the maximum of the pair-wise squared distances between vertices of sets ${\cal C}(\hat{\bm{s}}_i, \hat{\bm{R}}_i, \hat{\bm{l}}_i)$ and ${\cal C}(\bar{\bm{s}}_j, \bar{\bm{R}}_j, \bar{\bm{l}}_j)$.
\begin{align*}
 c^L_{i,j}=&\min \norm{\hat{p}_i-\bar{p}_j}_2^2 \\
& \hat{p}_i \in \textrm{conv}(\textrm{V}({\cal C}(\hat{\bm{s}}_i, \hat{\bm{R}}_i, \hat{\bm{l}}_i))) \label{CL} \tag{CL} \\
& \bar{p}_j \in \textrm{conv}(\textrm{V}({\cal C}(\bar{\bm{s}}_j, \bar{\bm{R}}_j, \bar{\bm{l}}_j))) 
\end{align*}
The model size reduction algorithm is summarized in \ref{algo:reduce}.
\begin{algorithm}
\caption{Model size reduction algorithm}
\label{algo:reduce}
\begin{algorithmic}[i]
\State Given $\alpha^{L},\alpha^{U},\beta^{L},\beta^{U},\gamma^{L},\gamma^{U}$ ${\cal B}$, ${\cal B}'={\cal B}$
\For{$(i,j) \in \cal B$} 
\State Compute $c^L_{i,j}$ and $c^U_{i,j}$ 
\If{$c^L_{i,j} > f^U$}
\State ${\cal B}'={\cal B}' \setminus (i,j)$
\EndIf
\If{$c^U_{i,j} > \min_{j: (i,j) \in \cal B} c^L_{i,j}$}
\State ${\cal B}'={\cal B}' \setminus (i,j)$
\EndIf
\EndFor
\State Return ${\cal B}'$ \label{state:ret}
\end{algorithmic}
\end{algorithm}\vspace*{-0.5cm}
\newpage
\section{Algorithm for the Global Optimum}
We solve problem \eqref{p:Align} to global optimality with a nested-spatial branch and bound (nsBB) approach. Problem \eqref{p:Align} with $|{\cal B}| \approx 10^7$ variables is intractable even for state-of-the-art MIQCQP solvers such as Gurobi~\cite{Gurobi}. However, in nsBB, due to our model-size reduction steps, we progressively create smaller problems that may then be solved with Gurobi. 

We maintain an outer-queue ${\cal Q}$, sorted by lower-bounds of each node. Each node in ${\cal Q}$ is branched on the three angles $\alpha, \beta, \gamma$ (by bisection) to create eight child nodes. For each child node $k$,
\begin{enumerate}
\item \textbf{Upper-bound.} Upper-bound $f^{U(k)}$ is computed by setting $\alpha^k, \beta^k, \gamma^k$ to the mid-points of $[\alpha^{L(k)},\alpha^{U(k)}]$, $[\beta^{L(k)},\beta^{U(k)}]$, $[\gamma^{L(k)},\gamma^{U(k)}]$ in \eqref{prob:ub}.
\begin{subequations}
\label{prob:ub}
\begin{align}
& \hat{p}^k_i = \hat{\bm{s}}_i    + \hat{\bm{R}}_i R_{\alpha^k\beta^k\gamma^k} \hat{\bm{l}}_i && \forall i \in  \hat{{\cal P}}  \label{ub:1}  \\ 
& \bar{p}^k_j = \bar{\bm{s}}_j   + \bar{\bm{R}}_j R_{\alpha^k\beta^k\gamma^k}\bar{\bm{l}}_j && \forall j \in  \bar{{\cal P}}  \label{ub:2}\\
& d^k_i=\min_{j \in  \bar{{\cal P}} }  \norm{(\hat{p}^k_i- \bar{p}^k_j)}_2^2 && \forall i \in  \hat{{\cal P}}\\
& f^{U(k)}=\sum_{i \in  \hat{{\cal P}} } d^k_i
\end{align}
\end{subequations}
If $f^{U(k)} \le f^U$, $f^U=f^{U(k)}$.
\item \textbf{Reduce.} We use Algorithm \ref{algo:reduce} with inputs $\alpha^{L(k)}$, $\alpha^{U(k)}$, $\beta^{L(k)}$, $\beta^{U(k)}$, $\gamma^{L(k)}$, $\gamma^{U(k)}$ to compute ${\cal B}^{'(k)}$. If ${\cal B}^{'(k)}= \emptyset$, go to step 4, else step 3.
\item \textbf{Lower-bound.} We solve problem $P^{(k)}$ \eqref{p:Align} using the reduced set ${\cal B}^{'(k)}$, with Gurobi. $f^{L(k)}$ is the lower-bound computed by Gurobi in time $t_{max}$.
\item \textbf{Prune.} If $f^{L(k)} > f^U$, or ${\cal B}^{'(k)}= \emptyset$ node $k$ is pruned, else it is added to ${\cal Q}$.
\end{enumerate}
$f^U$ is the upper-bound, $f^L$ (minimum of lower-bounds of all the nodes in ${\cal Q}$) is the lower bound. $g_r$ and $g_a$ are the relative and absolute gaps between the bounds, respectively. nsBB converges when either $g_r \le \epsilon_r$ or $g_a \le \epsilon_a$. nsBB is open-source, parallelized (multi-threaded and multi-node) written in C++, within the modelling platform Gravity \cite{Gravity}.
\subsection{Adaptive Grid Search}
The adaptive grid search (aGS) algorithm is used to generate feasible solutions fast that can be used as upper-bounds in nsBB. The aGS algorithm splits the domain of each angle into $n_d>0$ pieces, leading to a total of $n_d^3$ intervals. Within each interval, we solve problem \eqref{prob:ub}. We note the best-found upper-bound across the $n_d^3$ intervals $f^*$ and the corresponding solution $\alpha^*, \beta^*,\gamma^*$. We then set domain $[\alpha^{L},\alpha^{U}]$, $[\beta^{L},\beta^{U}]$, $[\gamma^{L},\gamma^{U}]$ to a perturbation of 10\% around $\alpha^*, \beta^*,\gamma^*$ and repeat the procedure until maximum time $T_{max}$ is reached. The aGS algorithm is multi-threaded.
\section{Experiments and Results}\label{sec:refs}
\begin{figure}
\begin{minipage}[b]{\linewidth}
  \centering
  \includegraphics[width=\textwidth]{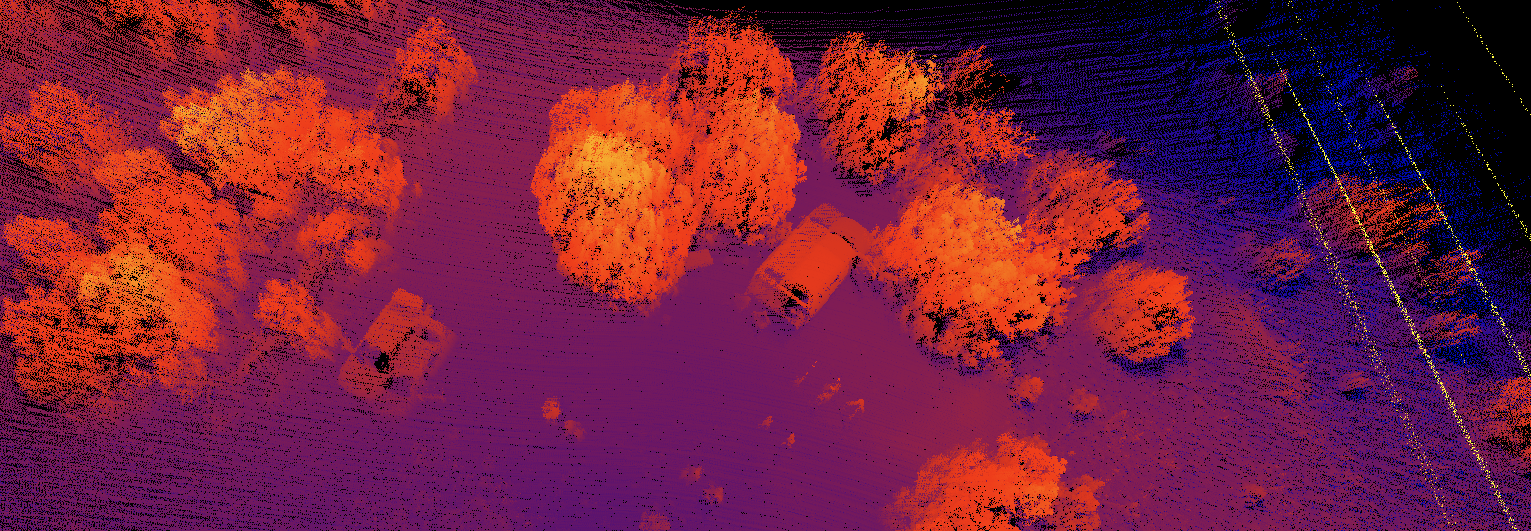}
\end{minipage}
\begin{minipage}[b]{\linewidth}
  \centering
  \includegraphics[width=\textwidth]{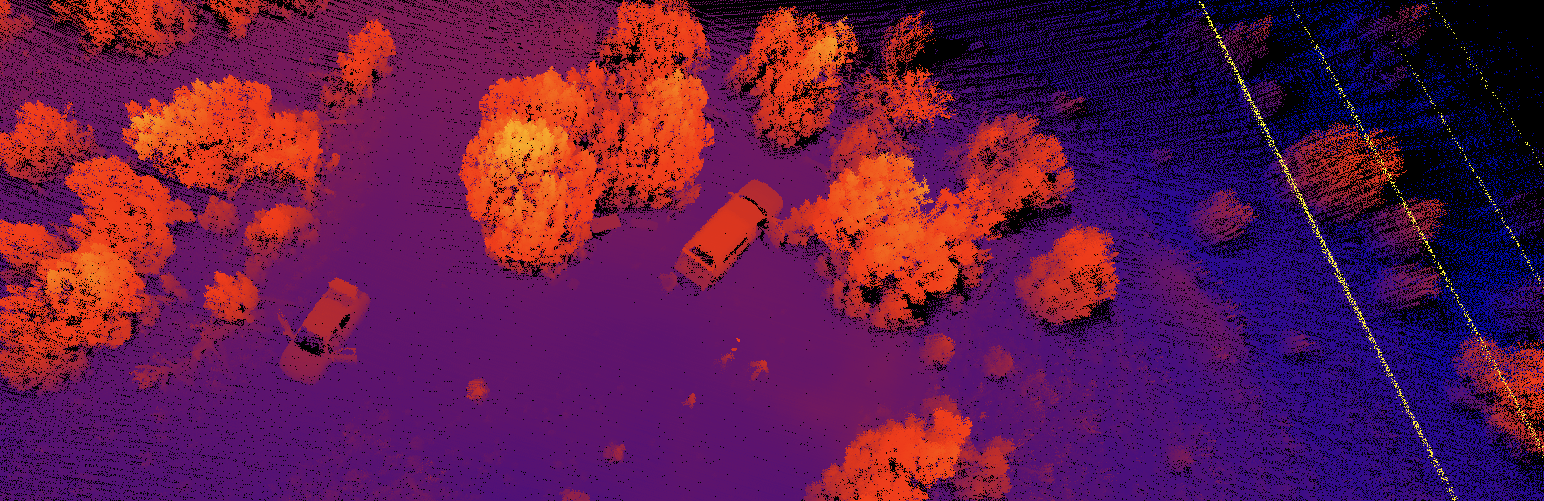}
\end{minipage}
\caption{Cars data set. Top: Initial. Below: After alignment.}
\label{fig:car}
\end{figure}
All algorithms and datasets presented in this work are open-source and available \href{https://github.com/lanl-ansi/ARMO}{here}.
\input{tables.tex}

In Table \ref{tab:data} we show details of three real-world data-sets. 
In Table \ref{tab:bb} we show the objective value, which is the total squared L2 error, and runtime in minutes, for aGS and nSBB algorithms. We use Intel Broadwell 2.10 GHz machines with 72 threads/node. We set $\alpha^L, \beta^L,\gamma^L$ to -2\degree and $\alpha^U, \beta^U,\gamma^U$ to 2\degree. The best-found solution by the aGS algorithm with $n_d=30$, $T_{max}=100$s, run on 1 node is shown. We use the aGS solution to initialize $f^U$ in the nSBB algorithm. The global optimum obtained with the nsBB algorithm with $\epsilon_r=0.01$, $\epsilon_a=0.1$, $t_{max}=30$s is shown next. Runtimes for nsBB are shown using 8 and 32 nodes. The solution-values (boresight angles) of the aCG and nsBB algorithms are given in Table \ref{tab:angles}. 
We solve MIQCQP \ref{p:Align} using Gurobi and nsBB on a subsampled car instance. Gurobi Cutoff is set to aGS $||L_2||^2$. As shown in Table \ref{tab:gurobi}, Gurobi alone makes no progress on \ref{p:Align}.

With the aGS algorithm we find high-quality solutions within 100 seconds. With the nsBB algorithm, using a small cluster of 8 nodes, we are able to identify optimal solutions and mathematically certify their global optimality for all instances in 1.5 to 4 hours. By using 32 nodes, nsBB converges in less than 90 minutes for all instances.

In Fig \ref{fig:car} (colorized by height) we plot $\cal \bar{{\cal D}}$ and $\hat{{\cal D}}$ for the car instance before and after applying the optimal boresight angles. 


\FloatBarrier
\section{Conclusion}
We present the first global optimization algorithm to solve the boresight alignment problem. We find high-quality heuristic solutions in seconds as well as the global optimum within hours. Our novel preprocessing steps, combined with parallelization and state-of-the-art solvers, enables solution of mixed-integer quadratic problems with greater than 20 million binary variables in under 90 minutes.

\vfill
\pagebreak

\bibliographystyle{IEEEbib}
{\small
\bibliography{Lidar.bib}}

\end{document}

%% file: tables.tex
\begin{table}[hbt!]
\vspace{-0.5cm}
\addtolength{\tabcolsep}{-1pt}
\begin{center}
\caption{
Data sets for boresight alignment
}
\label{tab:data}
\begin{tabular}{|l|c|c|c|c|c|}
\hline
Case & $|\bar{{\cal D}}|$ $(10^6)$ & $|\hat{{\cal D}}|$ $(10^6)$ & $|\bar{{\cal P}}|$ & $|\hat{{\cal P}}|$ & $|{\cal B}|$ $(10^6)$\\
\hline
Car & 2.59 & 1.86 & 9900 & 2075 & 20.54\\
Tent  & 5.24 & 5.19 & 8158 & 1052 & 8.58\\
Truck & 2.09 & 1.32 & 7766 & 1490 & 11.57\\
\hline
\end{tabular}
\end{center}
\vspace{-0.2cm}
\end{table}
\begin{table}[hbt!]
\addtolength{\tabcolsep}{-1.9pt}
\vspace{-0.5cm}
\begin{center}
\caption{
Performance of aGS and nsBB algorithms
}
\label{tab:bb}
\begin{tabular}{|l|c|c|c|c|c|c|} 
\hline
Case &	\multicolumn{3}{c|}{Objective $||L_2||^2$}							&	\multicolumn{3}{c|}{Time (min)}							\\
\hline
 & Initial & aGS & nsBB & aGS & nsBB-8 & nsBB-32 \\
Car & 873.5 & 12.4 & 11.9 & 1.7  & 187.6 & 78.3 \\
Tent & 12.1 & 1.3 &1.1 & 1.7 & 226.2 & 87.2 \\
Truck & 1870.5 & 8.0 & 7.9 & 1.7 & 89.7 & 39.1 \\
\hline
\end{tabular}
\end{center}
\vspace{-0.2cm}
\end{table}
\begin{table}[hbt!]
\vspace{-0.5cm}
\caption{Optimal boresight angles}
\addtolength{\tabcolsep}{-1.1pt}
\label{tab:angles}
\begin{center}
\begin{tabular}{|l|c|c|c|c|c|c|}
\hline
Case &	\multicolumn{6}{c|}{Angles (degrees)}\\
\hline
 &	\multicolumn{3}{c|}{aGS} & \multicolumn{3}{c|}{nsBB} \\
\hline
 & roll & pitch & yaw & roll & pitch & yaw \\
Car & -1.399 & 0.866 & -0.200 & -1.434 & 0.940 & -0.282\\
Tent & 0.072 & 0.626 & -0.218 & 0.126 & 0.729 & -0.325 \\
Truck & -1.538 & 0.852 & -0.180 & -1.528 & 0.835 & -0.141 \\
\hline
\end{tabular}
\end{center}
\vspace{-0.6cm}
\end{table}
\begin{table}[hbt!]
\addtolength{\tabcolsep}{-1.1pt}
\vspace{-0.2cm}
\caption{Gurobi: $|\hat{{\cal P}}|=462$, $|\bar{{\cal P}}|=495$, $|\cal{B}|$=$0.2\times10^5$}
\label{tab:gurobi}
\begin{center}
\begin{tabular}{|l|c|c|c|c|}
\hline
Method & $f^U$ & $f^L$ & $f^U-f^L$ & Time(h)\\
\hline
     nsBB-1 & 9.8 & 9.7& 0.1&  1.2\\
     Gurobi-1 & - & 0&- &  4 \\
     \hline
\end{tabular}
\end{center}
\vspace{-0.6cm}
\end{table}